\documentclass{article} 
\usepackage{iclr2023_conference,times}

\usepackage{amsmath,amsfonts,bm}

\def\eqref#1{equation~\ref{#1}}

\def\1{\bm{1}}

\DeclareMathAlphabet{\mathsfit}{\encodingdefault}{\sfdefault}{m}{sl}
\SetMathAlphabet{\mathsfit}{bold}{\encodingdefault}{\sfdefault}{bx}{n}

\usepackage{graphicx}
\usepackage{float}
\usepackage{subfigure}
\usepackage[colorlinks,linkcolor=pink]{hyperref}
\usepackage{url}
\usepackage{booktabs}
\usepackage{enumerate}
\usepackage{amssymb}
 
\title{Cross-level distillation and feature denoising for cross-domain few-shot classification}

\author{Hao ZHENG\footnotemark[1]  \\
Tokyo Institute of Technology\\
\texttt{zheng.h.ad@m.titech.ac.jp} \\ 
\And
Runqi Wang\footnotemark[1]\thanks{Co-first author.} \\
Huawei Noah's Ark Lab\\
\texttt{runqiwangstu@hotmail.com} \\
\And
Jianzhuang Liu \\
Huawei Noah's Ark Lab\\
\texttt{liu.jianzhuang@huawei.com} \\
\And
\ \ \ \ \ \ \ \ \ \ \ \ \ \ \ \ \ \ \ \ \ \ \ \ \  \ \ \ \ Asako Kanezaki\footnotemark[2]\thanks{Corresponding author.} \\
\ \ \ \ \ \ \ \ \ \ \ \ \ \ \ \ \ \ \ \ \ \ \ \ \  \ \ \ \ Tokyo Institute of Technology \\
\ \ \ \ \ \ \ \ \ \ \ \ \ \ \ \ \ \ \ \ \ \ \ \ \  \ \ \ \ \texttt{kanezaki@c.titech.ac.jp}
}

\iclrfinalcopy 
\begin{document}

\maketitle

\begin{abstract}
The conventional few-shot classification aims at learning a model on a large labeled base dataset and rapidly adapting to a target dataset that is from the same distribution as the base dataset. However, in practice, the base and the target datasets of few-shot classification are usually from different domains, which is the problem of cross-domain few-shot classification. We tackle this problem by making a small proportion of unlabeled images in the target domain accessible in the training stage. In this setup, even though the base data are sufficient and labeled, the large domain shift still makes transferring the knowledge from the base dataset difficult. We meticulously design a cross-level knowledge distillation method, which can strengthen the ability of the model to extract more discriminative features in the target dataset by guiding the network's shallow layers to learn higher-level information. Furthermore, in order to alleviate the overfitting in the evaluation stage, we propose a feature denoising operation which can reduce the feature redundancy and mitigate overfitting. Our approach can surpass the previous state-of-the-art method, Dynamic-Distillation, by 5.44$\%$ on 1-shot and 1.37$\%$ on 5-shot classification tasks on average in the BSCD-FSL benchmark. The implementation code will be available at \href{https://github.com/jarucezh/cldfd}{https://github.com/jarucezh/cldfd}.

\end{abstract}

\section{Introduction}
Deep learning has achieved great success on image recognition tasks with the help of a large number of labeled images. However, it is the exact opposite of the human perception mechanism which can recognize a new category by learning only a few samples. Besides, a large amount of annotations is costly and unavailable for some scenarios. 
It is more valuable to study few-shot classification which trains a classification model on a base dataset and rapidly adapts it to the target dataset. However, due to the constraint that the base data and the target data need to be consistent in their distributions, the conventional few-shot classification may not cater to the demands in some practical scenarios. For example, it may fail in scenarios where the training domain is natural images, but the evaluation domain is satellite images. Considering this domain shift in practical applications, we focus on cross-domain few-shot classification (CD-FSC) in this paper. Previous methods, such as ~\citep{mangla2020charting, adler2020cross, tseng2020cross}, can handle this problem with small domain gaps. However, the CD-FSC problem with a large domain gap is still a challenge.

BSCD-FSC \citep{guo2020broader} is a suitable benchmark for studying this problem, where the base dataset has natural images and the target datasets contain satellite images, crop disease images, skin disease images and X-ray images of sundry lung diseases. On this benchmark, previous methods following the traditional CD-FSC protocol train their models on the base dataset and evaluate them on the target dataset, but their performances are far from satisfactory. STARTUP \citep{phoo2020self} and Dynamic-Distillation \citep{islam2021dynamic} introduce a more realistic setup that makes a small portion of the unlabeled target images accessible during the training phase. These target images bring a prior to the model and dramatically promote the model's performance on the target datasets. Inspired by that, we follow their setup to explore the CD-FSC problem with a large domain shift.
\begin{figure}
\centering 
\includegraphics[width=0.85\textwidth]{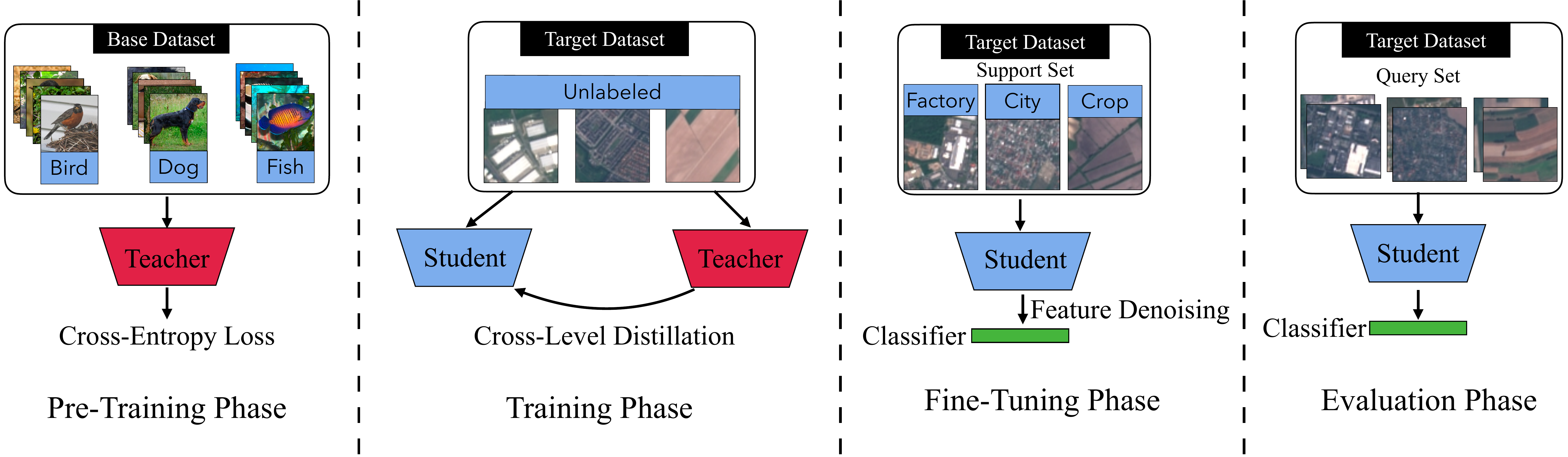} 
\caption{Our CD-FSC framework. The first phase is pre-training, which trains the teacher network on the labeled base dataset by optimizing the cross-entropy loss. The second phase trains the student network using our proposed cross-level distillation (CLD). The third phase fine-tunes a linear classifier on a few labeled images in the target domain, and feature denoising (FD) is conducted to remove the noise in the final feature vectors. The final phase classifies images in the target domain.} 
\label{fig_pipeline}
\end{figure}
In this work, we propose a cross-level distillation (CLD), which can effectively transfer the knowledge from the base dataset and improve the performance of the student network on the target domain. Besides, we propose feature denoising (FD) to remove the noise in the features during the fine-tuning stage. Our CD-FSC framework is given in Figure \ref{fig_pipeline}. 

The detail of CLD is shown in Figure \ref{fig_dis_structure}, which distills a teacher's deeper layers to a student's shallower layers, where the student and the teacher share the same structure. Unlike the distillation methods in STARTUP and Dynamic-Distillation, which only distill the teacher's last layer to the student's last layer, our CLD leads the shallow layers of the student to mimic the features generated from the deeper levels of the teacher so that the student can learn more deeper semantic information and extract more discriminative features on the target dataset. Additionally, since the teacher networks in STARTUP and Dynamic-Distillation are pre-trained on the base dataset only, the teacher's observation of the target data is biased. In order to calibrate the bias, we design an iterative process by building another network which shares the same structure and parameters with the historical student network, named old student network. In each training iteration, the features from the teacher and the old student in the same layers are dynamically fused to guide the corresponding layers of the student. The latter the training iteration, the fewer fusion features from the teacher network, and the more from the old student network. Due to the target data used in training is unlabeled, the self-supervised loss is introduced to excavate the target domain information further.

The self-supervised loss not only supports the network in mining valuable information on the target domain, but also brings a phenomenon where the final feature vector for classification has a small number of dominant (strongly activated) elements with the others are close to zero \citep{hua2021feature,kalibhat2022towards}. We find that during the fine-tuning phase in Figure~\ref{fig_pipeline}, these small activated elements are redundant and considered as noise. Our FD operation keeps the top $h$ largest elements and sets the others to zero. It is experimentally verified that FD can greatly improve the model's performance.

Above all, our main contributions are summarized below:
\begin{itemize}
    \item We propose a cross-level distillation (CLD) framework, which can well transfer the knowledge of the teacher trained on the base dataset to the student. We also use an old student network mechanism is also necessary to calibrate the teacher's bias learned from the base data.
    \item Considering the noisy feature activations, we design a feature denoising (FD) operation that can significantly improve the performance of our model.
    \item Extensive experiments are conducted to verify that our proposed CLD and FD can achieve state-of-the-art results on the BSCD-FSL benchmark with large domain gaps.
\end{itemize}

\section{Related Work}
\textbf{Cross-domain few-shot classification.} The cross-domain few-shot classification is firstly defined by \citep{chen2019closer}, which trains a model on the base dataset and evaluates it on the target dataset in a different domain from the base dataset. LFT \citep{tseng2020cross} simulates the domain shift from the base dataset to the target dataset by meta-learning and inserts linear layers into the network to align the features from the different domains. Meta-FDMixup \citep{fu2021meta} uses several labeled images from the target dataset for domain-shared feature disentanglement and feeds the domain-shared features to the classifier. FLUTE \citep{triantafillou2021learning} learns the universal templates of features across multi-source domains to improve the transferability of the model. However, all these methods concentrate on the CD-FSC problem with small domain shifts. 
Some methods handle CD-FSC with large domain gaps, in which the target datasets have obvious dissimilarity from the base dataset on perspective distortion, semantics, and/or color depth. For example, ConfeSS \citep{das2021confess} extracts useful feature components from labeled target images. ATA \citep{wang2021cross} does not require any prior of the target dataset and proposes a plug-and-play inductive bias-adaptive task augmentation module. CI \citep{luo2022channel} trains an encoder on the base dataset and converts the features of the target data with a transformation function in the evaluation stage. UniSiam \citep{lu2022self} adopts a self-supervised approach to address the CD-FSC problem. Among the methods dealing with large domain shifts, STARTUP \citep{phoo2020self} is a strong baseline, which uses a few unlabeled target images in the training stage. It firstly trains a teacher network on the base dataset in a supervised fashion and transfers the teacher's knowledge to the student by knowledge distillation (KD). It jointly optimizes the cross-entropy loss with the base dataset, contrastive loss of the unlabeled target images and the KD loss to upgrade the student network. Dynamic-Distillation \citep{islam2021dynamic} also uses a small number of unlabeled images and KD. The main difference between Dynamic-Distillation and STARTUP is that the former upgrades the pre-trained teacher dynamically by exponential moving averages, while the latter fixes the teacher. In our work, we follow their data setup allowing a small proportion of the unlabeled target images to be seen during the training phase. Different from the two methods that perform KD at the last layers of the teacher and the student, our KD is carried out at cross levels. Besides, our denoising operation further improves the performance.

\textbf{Self-supervised learning.} Self-supervised learning is widely used in the scenarios where labels are not available for training. It defines a ``pretext task'' to pre-train a network. For example, \citep{gidaris2018unsupervised} pre-trains the model by predicting the rotation angle of the image. One popular method is contrastive learning, such as SimCLR \citep{chen2020simple} which pulls different augmented versions of the same image closer and pushes the versions from different images away. Beyond contrastive learning, BYOL \citep{grill2020bootstrap} and SimSiam~\citep{chen2021exploring} rely only positive pairs.


\textbf{Knowledge distillation.} \citep{hinton2015distilling} firstly propose knowledge distillation by guiding a compact network (student) to mimic the output of a large network (teacher). Since features in the intermediate layers are informative, some previous methods distill the teacher's intermediate features \citep{romero2014fitnets} or attention maps of the features \citep{zagoruyko2016paying}. Besides, self-distillation methods, such as BYOT \citep{zhang2019your}, distill the last layer of the network to its own shallower layers. BYOT's teacher and student are the same network.

\begin{figure}
\centering 
\includegraphics[width=0.85\textwidth]{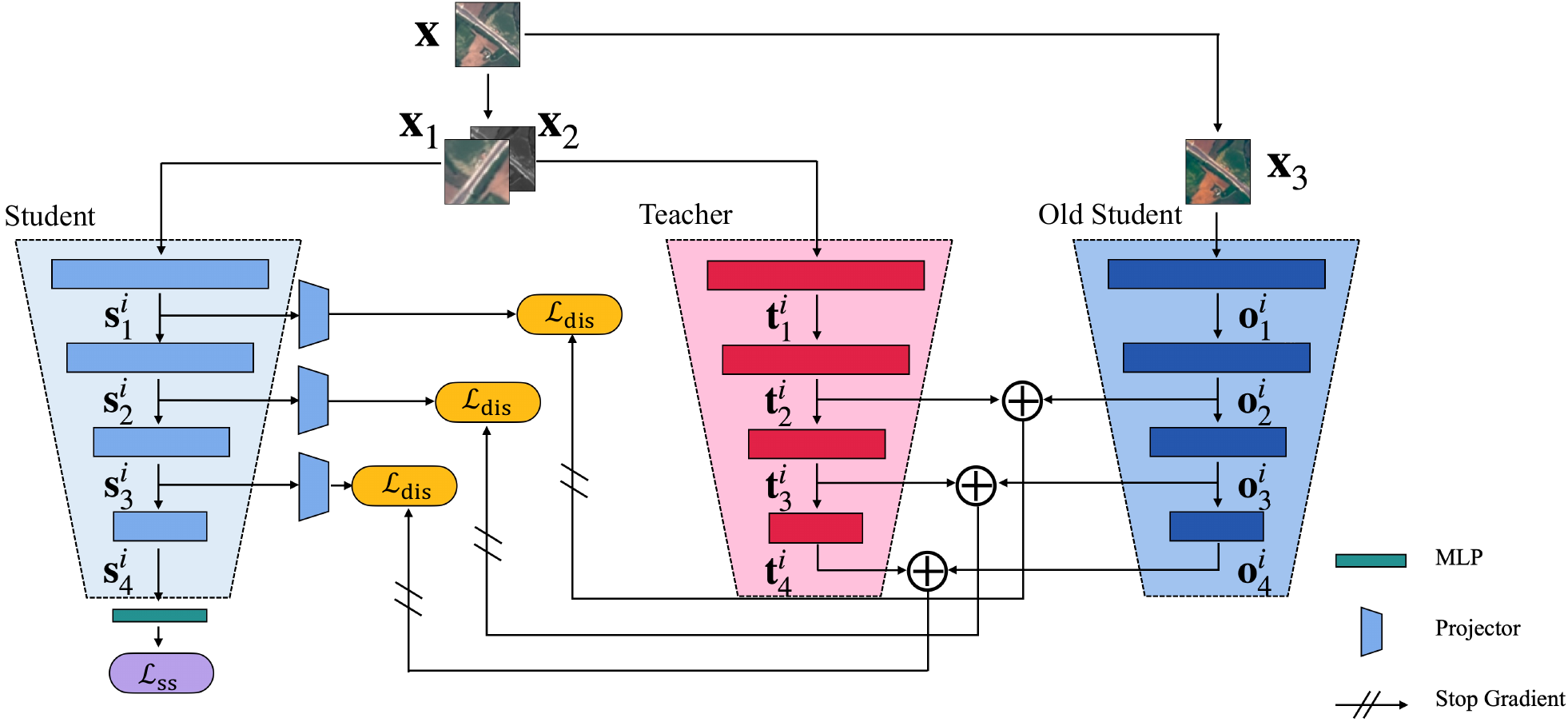} 
\caption{Framework of CLD for knowledge distillation (KD). The teacher network is pre-trained on the labeled base dataset with the cross-entropy loss and is fixed during KD. When training the student network, the target image $\mathbf{x}$ is augmented into $\mathbf{x}_{1}$, $\mathbf{x}_{2}$ and $\mathbf{x}_{3}$. Then ${\mathbf{x}_{1}, \mathbf{x}_{2}}$ is fed to both the student and the teacher, and $\mathbf{x}_{3}$ is fed to the old student. At the $i$-th iteration, the parameters of the old student are a copy of those of the student at $(i-\tau)$-th iteration. The feature $\mathbf{s}_{l}^i$ of the student is firstly projected by $\omega_{l}$ for dimensionality alignment, where $l$ is the block index. Then we fuse the features $\mathbf{t}_{l+1}^i$ and $\mathbf{o}_{l+1}^i$ obtaining $\mathbf{u}_{l+1}^i$, which are from the $(l+1)$-th block of the teacher and the old student, respectively. The KD is conducted by forcing $\omega_{l}(\mathbf{s}_{l}^i)$ to mimic $\mathbf{u}_{l+1}^i$. Additionally, the self-supervised loss $\mathcal{L}_{ss}$ is introduced on the student network.}
\label{fig_dis_structure} 
\end{figure}

In our framework (Figure~\ref{fig_dis_structure}), in addition to the KD in the intermediate layers, we design an old student that shares the same structure as the student but with different parameters. The introduction of this old student not only alleviates the teacher's bias learned from the base dataset, but also has the effect of assembling multiple historic students during training.

\section{Methodology}

\subsection{Preliminary}
\label{sec31_preliminary}

During the training period, we follow the setting in STARTUP \citep{phoo2020self} and Dynamic-Distillation \citep{islam2021dynamic}, 
where a labeled base dataset $\mathcal{D}_{\rm B}$ and a few unlabeled target images sampled from the target dataset $\mathcal{D}_{\rm T}$ are available.
In the testing stage, the support set $\mathcal{D}_{\rm S}$ comprises $N$ classes, and $K$ samples are randomly selected from each class in $\mathcal{D}_{\rm T}$, which is the so-called $N$-way $K$-shot task. The support set $\mathcal{D}_{\rm S}$ is for fine-tuning a new classifier with the frozen encoder (the student network in this work). The images in the query set $\mathcal{D}_{\rm Q}$ are randomly picked from the selected $N$ classes for evaluating the classification accuracy. The support set $\mathcal{D}_{\rm S}$ and the query set $\mathcal{D}_{\rm Q}$ have no overlap.

\subsection{Cross-level distillation}
\label{sec32_crosslevel}

The proposed cross-level distillation (CLD) framework is shown in Figure~\ref{fig_dis_structure}. The teacher network $f_{\rm t}$ is pre-trained on $\mathcal{D}_{\rm B}$ with the cross-entropy loss. The student network $f_{\rm s}$ is expected to inherit the knowledge of the teacher and extract discriminative features of $\mathcal{D}_{\rm T}$. However, the teacher's observation of the target data is biased since it is pre-trained on the base dataset only. In the $i$-th training iteration, if the features extracted by the student $f_{\rm s}^{i}$ directly mimic the features of the teacher, the teacher's bias will be transferred to the student. To reduce the bias, we introduce an \emph{old student network} $f_{\rm o}^{i}$, which is a copy of $f_{\rm s}^{i-\tau}$, where the hyper-parameter $\tau$ denotes the training iteration interval between $f_{\rm s}^{i-\tau}$ and $f_{\rm s}^{i}$.

To simplify the KD complexity, we divide each backbone of $f_{\rm s}$, $f_{\rm o}$, and $f_{\rm t}$ into $L$ residual blocks. Let $\mathbf{s}_l^i$, $\mathbf{o}_l^i$, and $\mathbf{t}_l^i$ be the features obtained by the student, the old student, and the teacher in the $l$-th block at the $i$-th iteration. The fusion between $\mathbf{t}_l^i$ and $\mathbf{o}_l^i$ is defined as:
\begin{equation}
   \mathbf{u}_l^i  = \alpha^i \mathbf{o}_l^i + (1-\alpha^i)\mathbf{t}_l^i,
\label{equ:united_teacher}
\end{equation}

where $\alpha^i=\frac{i}{T}$ is a dynamic weight with $T$ being the total number of training iterations. The KD loss $\mathcal{L}_{\rm {dis}}^i$ in the $i$-th iteration is defined as:
\begin{equation}
    \mathcal{L}_{\rm {dis}}^{i} = \begin{cases} \sum_{l=1}^{L}\Vert \omega_l(\mathbf{s}_l^{i}) - \mathbf{t}_{l+1}^{i} \Vert_2^2 & \text{if}\ i \leq \tau\\ \\
    \sum_{l=1}^{L}\Vert \omega_l(\mathbf{s}_l^{i}) - \mathbf{u}_{l+1}^{i} \Vert_2^2 & \text{otherwise,} \end{cases}
\label{equ:loss_dis}
\end{equation}
where $\omega_l(\cdot)$ is a projector of the $l$-th student's block for feature dimensionality alignment. It is comprised of a convolutional layer, a batch normalization operator and ReLU activation. Note that the KD in Equation \ref{equ:loss_dis} is from the $(l+1)$-th block of the teacher to the $l$-th block of the student. In our experiments, we find that this style is better than others (see Section \ref{ablation_cld}).

The total loss $\mathcal{L}$ for training the student network is:
\begin{equation}
    \mathcal{L} = \mathcal{L}_{\rm {ss}} + \lambda \mathcal{L}_{\rm {dis}},
\label{equ:loss_total}
\end{equation}
where $\mathcal{L}_{\rm {ss}}$ is a self-supervised loss, and $\lambda$ is a weight coefficient of loss function for balancing the two losses.

For $\mathcal{L}_{\rm {ss}}$, off-the-shelf self-supervised losses like SimCLR~\citep{chen2020simple} or BYOL~\citep{grill2020bootstrap} can be used. The contrastive loss in SimCLR is:
\begin{equation}
    \mathcal{L}_{\rm{simclr}}=-\frac{1}{\left| \mathbf{B} \right|} \sum_{m, n \in \mathbf{B}} \log \frac{\exp \left(\operatorname{sim}\left(\mathbf{z}_m, \mathbf{z}_n\right) / \gamma \right)}{\sum_{q=1, q \neq m}^{2 \left| \mathbf{B} \right|}  \exp \left(\operatorname{sim}\left(\mathbf{z}_m, \mathbf{z}_q\right) / \gamma\right)},
    \label{eq:simclr}
\end{equation}

where $\mathbf{z}_m$, $\mathbf{z}_n$ and $\mathbf{z}_q$ are the projected embeddings of different augmentations, $(m, n)$ is a positive pair from the same image, $\operatorname{sim}(\cdot)$ is a similarity function, $\mathbf{B}$ is a mini-batch of unlabeled target images, and $\gamma$ is a temperature coefficient. The self-supervised loss in BYOL is:
\begin{equation}
    \mathcal{L}_{\rm{byol}}=\sum_{m \in \mathbf{B}}2 - 2\cdot\frac{\operatorname{sim}(p(\mathbf{z}_m), \mathbf{z}_m^{\prime})}{ \left\|p(\mathbf{z}_m)\right\|_2 \cdot \left\|\mathbf{z}_m{\prime}\right\|_2},
    \label{eq:byol}
\end{equation}

where $\mathbf{z}_m$ and $\mathbf{z}_m^{\prime}$ are embeddings of the online network and the target network, respectively, and $p(\cdot)$ is a linear predictor.

Note that the last convolution block of the student network is not involved in KD and is trained by minimizing $\mathcal{L}_{\rm{ss}}$ only. The reason is that the last block mainly discovers semantic information that is highly domain-specific. Therefore, we constrain it on the target data rather than letting it learn from the teacher that is pre-trained on the base data.

\subsection{Feature Denoising}
\label{sec_featuredenoise}

The self-supervised loss brings a phenomenon where the final feature vector for classification has a small number of strongly activated elements while the others are close to zero \citep{hua2021feature, kalibhat2022towards}. These elements of small magnitudes are regarded as noise, which may cause overfitting. We propose a feature denoising (FD) operation to 
remove their contribution during the fine-tuning phase (see Figure~\ref{fig_pipeline}). FD keeps the largest $h$ elements of the feature from the student network and zeros the other elements. The FD operation is only performed on the feature $\mathbf{s}_{L}$ in the last layer of the student network. Specifically, let $\mathbf{s}_{L} = [s_{(L, 1)}, s_{(L, 2)},\dots,s_{(L, D_L)}]$ and $\widetilde{\mathbf{s}}_{L} = [\widetilde{s}_{(L, 1)}, \widetilde{s}_{(L, 2)},\dots,\widetilde{s}_{(L, D_L)}]$ be the features before and after the FD operation, respectively, where $D_L$ is the feature's dimensionality. Then the FD operation is defined as:
\begin{equation}
    \widetilde{s}_{L,d} = \begin{cases} ({s_{L, d}})^\beta & \text{ if }\ {s_{L, d}} \in top_h(\mathbf{s}_{L}),\ d=1, 2, ..., D_L \\
    0 & \text{ otherwise},
    \end{cases}
    \label{equ_fd}
\end{equation}

where $\beta$ is a hyper-parameter which makes the non-zero elements more distinguishable and $top_h(\cdot)$ is the operator selecting the largest $h$ elements of the feature. Finally, $\widetilde{\mathbf{s}}_{M}$ is fed to the classifier for fine-tuning.

\section{Experiments}
\subsection{Datasets and Implementation}
\textbf{Datasets.} We evaluate the proposed CLD and FD for the CD-FSC problem on the BSCD-FSL benchmark~\citep{guo2020broader} with large domain gaps. The miniImageNet dataset~\citep{vinyals2016matching} serves  as the base dataset $\mathcal{D}_B$ which has sufficient labeled images. EuroSAT \citep{helber2019eurosat}, CropDisease \citep{mohanty2016using}, ISIC \citep{codella2019skin} and ChestX \citep{wang2017chestx} in BSCD-FSL are the unlabeled target datasets $\mathcal{D}_T$. We follow the training protocol in STARTUP ~\citep{phoo2020self} and Dynamic-Distillation~\citep{islam2021dynamic}, which allows the whole labeled training set of miniImageNet and a small proportion (20$\%$) of unlabeled target images available during the training period. The remaining 80$\%$ of target images are utilized for fine-tuning and evaluating by building $5$-way $K$-shot tasks, $K\in\{1, 5\}$.

\textbf{Implementation details.} For a fair comparison, all the methods use ResNet-10 ~\citep{guo2020broader} as the backbone. Our model is optimized by SGD with the momentum 0.9, weight decay 1e-4, and batch size 32 for 600 epochs. The learning rate is 0.1 at the beginning and  decays by $0.1$ after the $300$th epoch and $500$th epoch.
The hyper-parameters $\lambda$ in Equation \ref{equ:loss_total}, and $h$ and $\beta$ in Equation~\ref{equ_fd} are set to $2$, $64$, and $0.4$, respectively. For fine-tuning and evaluating the student network, we randomly sample  $600$ episodes of $5$-way $K$-shot tasks. The performance is represented by the average classification accuracy over the 600 episodes within the $95\%$ confidence interval. In each episode, the parameters of the student are frozen. An additional linear classifier is fine-tuned by minimizing the cross-entropy loss. The above mentioned hyperparameters are determined based on the EuroSAT dataset only, and then they are used for evaluation on all the datasets, showing the generalization ability of our method.

\renewcommand{\arraystretch}{1.5}
\begin{table}[t]
\caption{The averaged 5-way 1-shot and 5-shot averaged accuracy and 95\% confidence interval among 600 episodes are given. The reported results of SimCLR (Base) and previous state-of-the-art Dynamic-Distillation are from \citep{islam2021dynamic}. The results of CI are from \citep{luo2022channel}. The results of ATA are from \citep{wang2021cross}. The performance of ConFeSS refers to \citep{das2021confess} and they do not give the confidence interval. The champion results are marked in bold.}
\label{tab_mainresults}
\resizebox{1\columnwidth}{!}
{
\begin{tabular}{@{}ccccccccc@{}}
    
\toprule
& \multicolumn{2}{c}{EuroSAT} & \multicolumn{2}{c}{CropDisease} & \multicolumn{2}{c}{ISIC} & \multicolumn{2}{c}{ChestX} \\ \cmidrule(lr){2-3}\cmidrule(lr){4-5}\cmidrule(lr){6-7}\cmidrule(lr){8-9} 
& \multicolumn{1}{c}{1-shot} & \multicolumn{1}{c}{5-shot} & \multicolumn{1}{c}{1-shot} & \multicolumn{1}{c}{5-shot} & \multicolumn{1}{c}{1-shot} & \multicolumn{1}{c}{5-shot} & \multicolumn{1}{c}{1-shot}       & \multicolumn{1}{c}{5-shot}      \\ \midrule
\multicolumn{1}{c}{Transfer}&58.42$\pm$0.94 &75.31$\pm$0.71 &68.45$\pm$0.87        &89.12$\pm$0.52 &32.82$\pm$0.60 &47.13$\pm$0.58 &22.48$\pm$0.41 &26.65$\pm$0.43 \\
\multicolumn{1}{c}{SimCLR (Base)}&58.28$\pm$0.90 &80.83$\pm$0.64 &68.26$\pm$0.86 &83.44$\pm$0.61 &32.15$\pm$0.59 &45.90$\pm$0.58 &22.37$\pm$0.42 &26.63$\pm$0.46 \\
\multicolumn{1}{c}{CI}&58.82$\pm$0.92 &76.26$\pm$0.70 &61.58$\pm$0.88 &89.25$\pm$0.51 &32.43$\pm$0.56 &44.04$\pm$0.55 &23.23$\pm$0.41 &27.20$\pm$0.44 \\
\multicolumn{1}{c}{Transfer+SimCLR}&65.92$\pm$0.88 &81.83$\pm$0.59 &81.39$\pm$0.80 &94.85$\pm$0.38 &33.86$\pm$0.61 &47.25$\pm$0.59 &22.91$\pm$0.45 &27.03$\pm$0.43 \\
\multicolumn{1}{c}{STARTUP}&65.24$\pm$0.88 &81.60$\pm$0.59 &78.18$\pm$0.83 &92.86$\pm$0.44 &34.15$\pm$0.62 &46.42$\pm$0.58 &22.86$\pm$0.43 &26.98$\pm$0.43 \\
\multicolumn{1}{c}{ATA} &65.94$\pm$0.50 &79.47$\pm$0.30 &77.82$\pm$0.50 &88.15$\pm$0.50 &34.70$\pm$0.40 &45.83$\pm$0.30 &21.67$\pm$0.20 &23.60$\pm$0.20  \\
\multicolumn{1}{c}{BYOL} &64.38$\pm$0.93 &82.44$\pm$0.56 &67.15$\pm$1.17 &89.40$\pm$0.63 &29.21$\pm$0.51 &41.44$\pm$0.51 &21.66$\pm$0.31 &25.99$\pm$0.43  \\
\multicolumn{1}{c}{ConFeSS} &$-$ &84.65 &$-$ &88.88 &$-$ &48.85 &$-$ &27.09  \\
\multicolumn{1}{c}{Dynamic-Distillation}&73.14$\pm$0.84 &89.07$\pm$0.47 &82.14$\pm$0.78 &95.54$\pm$0.38 &34.66$\pm$0.58 &49.36$\pm$0.59 &23.38$\pm$0.43 &28.31$\pm$0.46 \\
\multicolumn{1}{c}{SimCLR} &75.10$\pm$0.85 &90.17$\pm$0.43 &88.54$\pm$0.80 &96.09$\pm$0.39 &35.71$\pm$0.66 &48.84$\pm$0.61 &22.00$\pm$0.42 &24.84$\pm$0.41  \\
\hline
\multicolumn{1}{c}{BYOL+CLD+FD (ours)} &72.78$\pm$0.84 &88.50$\pm$0.45 &86.94$\pm$0.73 &95.85$\pm$0.38 &37.70$\pm$0.67 &51.19$\pm$0.63 &\textbf{24.34$\pm$0.44} &\textbf{30.15$\pm$0.44} \\
\multicolumn{1}{c}{SimCLR+CLD+FD (ours)} &\textbf{82.52$\pm$0.76} &\textbf{92.89$\pm$0.34} &\textbf{90.48$\pm$0.72} &\textbf{96.58$\pm$0.39} &\textbf{39.70$\pm$0.69} &\textbf{52.29$\pm$0.62} &22.39$\pm$0.44  &25.98$\pm$0.43 \\ \bottomrule
\end{tabular}}
\end{table}

\begin{figure}[t]
\centering 
\includegraphics[width=0.8\textwidth]{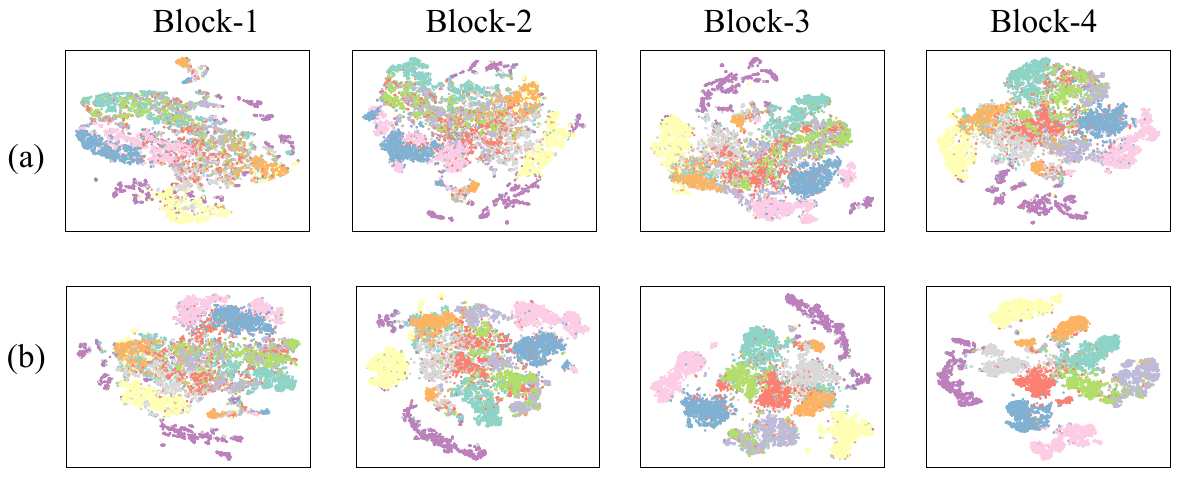} 
\caption{t-SNE results of (a) STARTUP and (b) SimCLR+CLD+FD at different blocks on the EuroSAT dataset. It is obvious that our method can extract more discriminative features.} 
\label{t-sne}
\end{figure}

\subsection{Main Results}
\label{sec_mainresults}
We select several basic and competitive methods for comparison in Table~\ref{tab_mainresults}. The basic model is Transfer that trains the encoder on the labeled base dataset with the cross-entropy loss. SimCLR (Base)~\citep{chen2020simple} trains the model on the base dataset (miniImageNet) with the contrastive loss of SimCLR. 
CI~\citep{luo2022channel} trains the encoder on the base dataset and converts the features of the target dataset with a transformation function in the evaluation stage. Transfer+SimCLR is proposed in \citep{islam2021broad} and exhibits good transferability, which simultaneously optimizes the cross-entropy loss on the base dataset and the contrastive loss of SimCLR on the target dataset. STARTUP~\citep{phoo2020self} trains the model with three loss functions: cross-entropy loss on the labeled base dataset, and contrastive loss and KD loss on the unlabeled target domain. ATA~\citep{wang2021cross} designs a plug-and-play inductive bias-adaptive task augmentation module. BYOL~\citep{grill2020bootstrap} is trained on the target dataset. ConfeSS~\citep{das2021confess} uses labeled target images to find out the useful components of the target image features. Dynamic-Distillation~\citep{islam2021dynamic} designs a distillation loss that draws support from the dynamically updated teacher network. SimCLR trains the encoder on unlabeled target data only and shows good generalization ability. For our method, we build two models BYOL+CLD+FD and SimCLR+CLD+FD, which use Equations \ref{eq:byol} and \ref{eq:simclr} for self-supervised losses, respectively.

\begin{figure}[t]
\centering 
\includegraphics[width=0.85\textwidth]{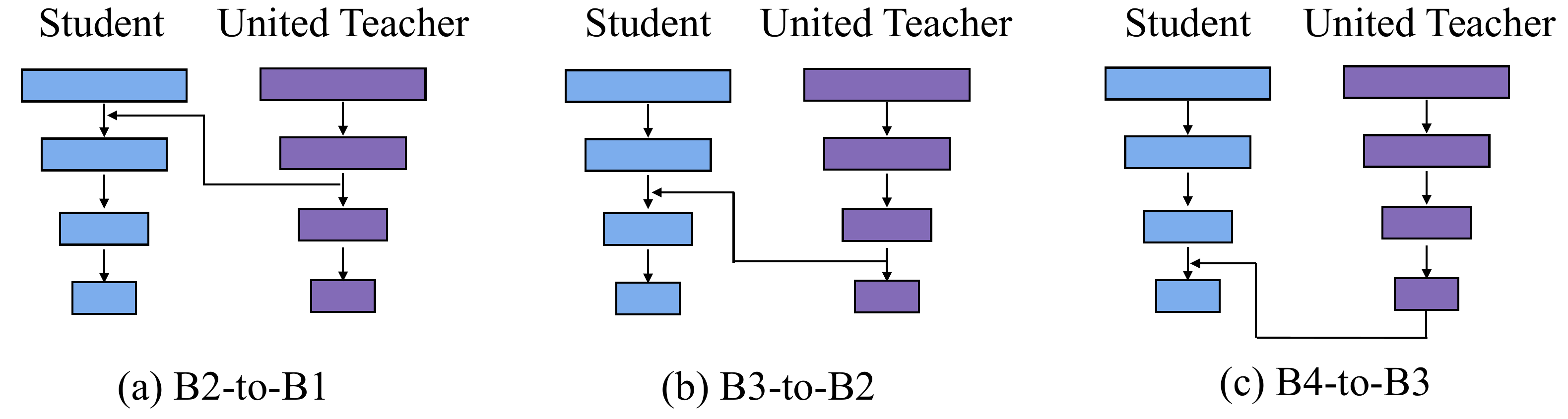}
\caption{Different single-block distillation structures. B2-to-B1 means that the second block of the teacher is distilled to the first block of the student. B3-to-B2 and B4-to-B3 are similar.} 
\label{single_level} 
\end{figure}

\begin{table}[t]
\center
\caption{Results of different single-block distillation stuctures on EuroSAT (5-way 1-shot). B2-to-B1, B3-to-B2, and B4-to-B3 are defined in Figure \ref{single_level}.}
\resizebox{0.7\columnwidth}{!}{
\begin{tabular}{cccccc}
\toprule
&SimCLR &B2-to-B1 &B3-to-B2 &B4-to-B3 &CLD 
 \\ \hline
&75.10$\pm$0.85 &78.54$\pm$0.80 &78.38$\pm$0.77 &77.92$\pm$0.83 &79.38$\pm$0.82
 \\ \bottomrule
\end{tabular}}
\label{tab_single}
\end{table}

Table~\ref{tab_mainresults} gives the comparisons among our models and the baselines. All the best results on the four datasets are obtained by either of our two models. In particular on EuroSAT, our SimCLR+CLD+FD can bring 9.38\% and 7.42\% gains over Dynamic-Distillation and SimCLR on the $5$-way $1$-shot task, respectively. On average, on the four datasets, SimCLD+CLD+FD outperforms Dynamic-Distillation significantly (58.77\% vs. 53.33\%) for 1-shot; 66.94\% vs. 65.57\% for 5-shot. Besides, although SimCLR+CLD+FD and BYOL+CLD+FD are based on SimCLR and BYOL, respectively, the performances of SimCLR and BYOL are improved greatly. Finally, we visualize the feature distributions of STARTUP and our SimCLR+CLD+FD in Figure \ref{t-sne}. It can be seen that SimCLR+CLD+FD exhibits better clustering results at all the blocks, especially at the last one.

\subsection{Ablation Study on Cross-level Distillation}
\label{ablation_cld}

\begin{figure}[hb]
\centering 
\includegraphics[width=0.85\textwidth]{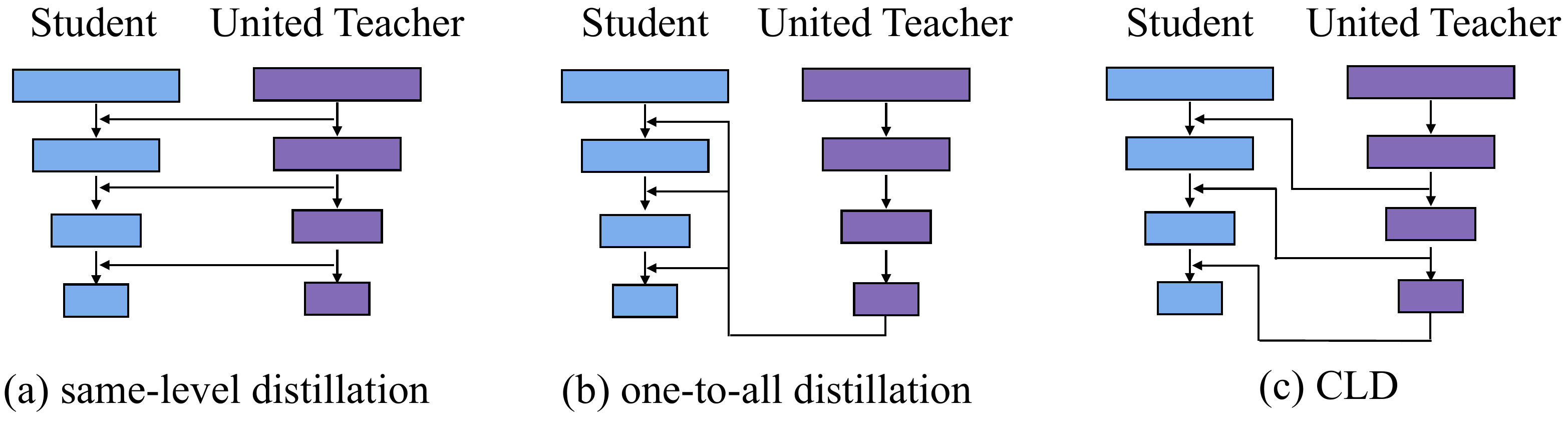} 
\caption{Different multi-blocks distillation structures. (a) and (b) are two common KD methods. (c) is our cross-level KD.} 
\label{cross_level} 
\end{figure}

\begin{table}[ht]
\caption{Results of different multi-blocks KD structures.}
\resizebox{1\columnwidth}{!}
{
\begin{tabular}{@{}ccccccccc@{}}
\toprule
                & \multicolumn{2}{c}{EuroSAT} & \multicolumn{2}{c}{CropDisease} & \multicolumn{2}{c}{ISIC} & \multicolumn{2}{c}{ChestX} \\
                & 1-shot       & 5-shot       & 1-shot         & 5-shot         & 1-shot      & 5-shot     & 1-shot       & 5-shot      \\ \midrule
SimCLR &75.10$\pm$0.85 &90.17$\pm$0.43 &88.54$\pm$0.80 &\textbf{96.09$\pm$0.39} &35.71$\pm$0.66 &48.84$\pm$0.61 &22.00$\pm$0.42 &24.84$\pm$0.41  \\
same-level distillation &75.48$\pm$0.85 &90.71$\pm$0.40 &88.14$\pm$0.80 &95.76$\pm$0.44 &36.56$\pm$0.65 &50.51$\pm$0.58 &\textbf{22.12$\pm$0.42} &\textbf{25.20$\pm$0.42}             \\
one-to-all distillation &77.70$\pm$0.83 &91.80$\pm$0.38 &89.11$\pm$0.79 &95.83$\pm$0.44 &36.69$\pm$0.64 &50.64$\pm$0.62  &22.10$\pm$0.43 &24.77$\pm$0.43             \\
CLD &\textbf{79.38$\pm$0.82} &\textbf{92.78$\pm$0.35} &\textbf{89.43$\pm$0.78} &96.08$\pm$0.43 &\textbf{37.82$\pm$0.68} &\textbf{52.28$\pm$0.61} &22.08$\pm$0.43 &25.08$\pm$0.42             \\ \bottomrule
\end{tabular}}
\label{tab_cross_stra}
\end{table}


\begin{figure}[ht]
\centering
\includegraphics[width=0.6\linewidth]{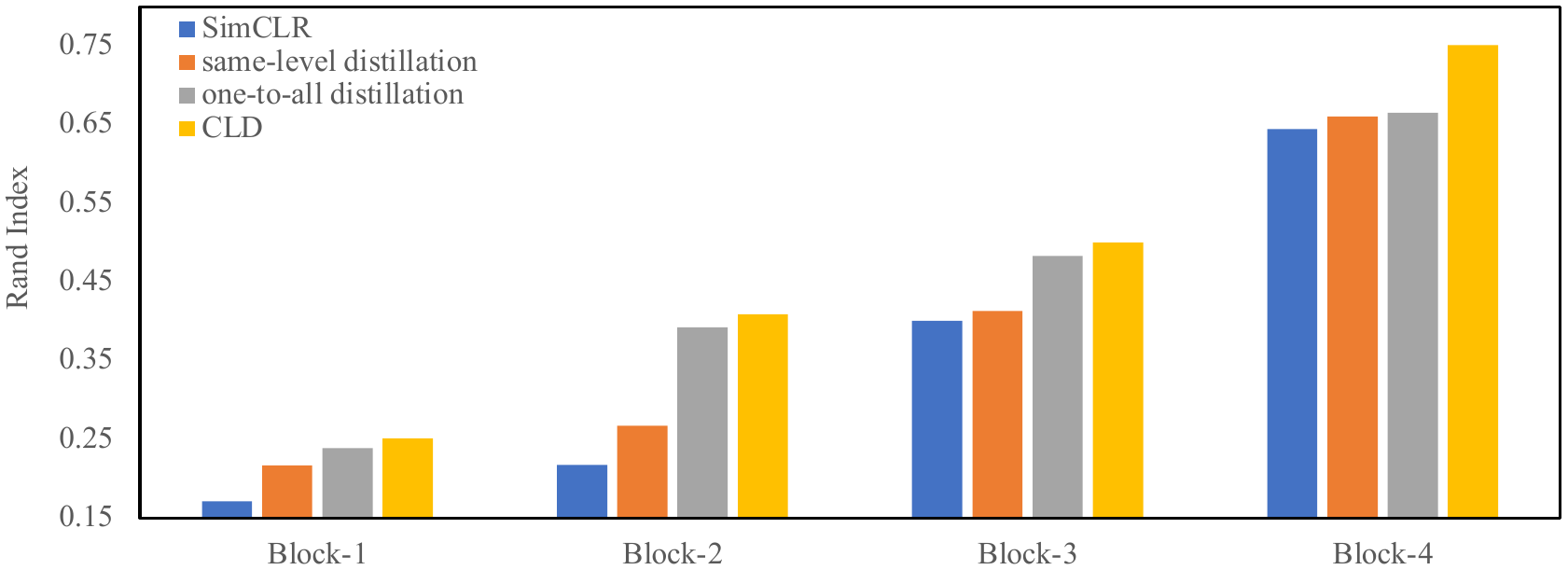}
\caption{Rand Index values in the residual blocks of the methods in Table \ref{tab_cross_stra}.} 
\label{fig_randindex} 
\end{figure}

For simplicity, the combination (according to Equation \ref{equ:united_teacher}) of the teacher and the old student in Figure \ref{fig_dis_structure} is denoted as the united teacher in Figure \ref{single_level} and Figure \ref{cross_level}. In these experiments, $\mathcal{L}_{\rm{simclr}}$ is used as $\mathcal{L}_{\rm{ss}}$ and FD is not employed. Table \ref{tab_single} gives the results of SimCLR, different single-block distillation structures, and CLD on EuroSAT (5-way 1-shot). We can notice that all the single-block distillation structures perform better than SimCLR, and CLD can outperform all the single-block distillation structures. Table \ref{tab_cross_stra} gives the results of SimCLR, different multi-blocks distillation structures and CLD. Compared with SimCLR without KD, all the multi-blocks KD structures improve the model's overall performance. Second, the one-to-all KD outperforms the same-level KD in most cases. Finally, our CLD performs best on average among the three multi-blocks KD structures.


We further explore the reason that CLD can exceed other methods. Rand Index \citep{rand1971objective} is a metric to reflect the quality of feature clustering. The bigger Rand Index is, the better clustering the method provides. Figure \ref{fig_randindex} is comprised of the Rand Index values of the features in each residual block on the EuroSAT test dataset. Our CLD increases more than all the other methods in each block. It means that our CLD can pick up more discriminative information at each level so that the model gradually hunts the useful features as the network deepens.


Next, we examine the usefulness of the old student and the effect of the training iteration interval $\tau$ (Section \ref{sec32_crosslevel}). Table \ref{tab_tau} shows the experimental results of different settings. First, CLD without the old student outperforms SimCLR. Second, using the old student is better than not using it. Considering the performance and the memory requirement, we choose $\tau=1$ in all the other experiments on the four datasets.

\begin{table}[h]
\caption{Effects of the old student and the training iteration interval $\tau$. All the methods in the table use $\mathcal{L}_{\rm{simclr}}$ as $\mathcal{L}_{\rm{ss}}$.}
\resizebox{1\columnwidth}{!}
{
\begin{tabular}{@{}ccccccccc@{}}
\toprule
                & \multicolumn{2}{c}{EuroSAT} & \multicolumn{2}{c}{CropDisease} & \multicolumn{2}{c}{ISIC} & \multicolumn{2}{c}{ChestX} \\
                & 1-shot       & 5-shot       & 1-shot         & 5-shot         & 1-shot      & 5-shot     & 1-shot       & 5-shot      \\ \midrule
SimCLR &75.10$\pm$0.85 &90.17$\pm$0.43 &88.54$\pm$0.80 &\textbf{96.09$\pm$0.39} &35.71$\pm$0.66 &48.84$\pm$0.61 &22.00$\pm$0.42 &24.84$\pm$0.41  \\
CLD (w/o old student) &76.85$\pm$0.83 &91.58$\pm$0.40 &88.72$\pm$0.78 &95.85$\pm$0.43 &37.04$\pm$0.65 &50.90$\pm$0.60 &21.40$\pm$0.32 &23.91$\pm$0.39             \\
CLD ($\tau=0$) &78.56$\pm$0.82 &92.32$\pm$0.35 &89.11$\pm$0.81 &95.69$\pm$0.45 &37.82$\pm$0.69 &51.17$\pm$0.62 &22.03$\pm$0.43 &25.12$\pm$0.44             \\
CLD ($\tau=1$) &\textbf{79.38$\pm$0.82} &\textbf{92.78$\pm$0.35} &89.43$\pm$0.78 &96.08$\pm$0.43 &37.82$\pm$0.68 &\textbf{52.28$\pm$0.61} &\textbf{22.08$\pm$0.43} &25.08$\pm$0.42\\
CLD ($\tau=10$) &78.18$\pm$0.81 &92.17$\pm$0.36 &89.24$\pm$0.80 &95.83$\pm$0.44 &\textbf{38.02$\pm$0.68} &51.77$\pm$0.62 &22.02$\pm$0.43 &25.04$\pm$0.43             \\
CLD ($\tau=100$) &78.79$\pm$0.80 &92.44$\pm$0.37 &\textbf{89.56$\pm$0.81} &96.06$\pm$0.42 &36.81$\pm$0.65 &50.60$\pm$0.62 &22.29$\pm$0.43 &\textbf{25.17$\pm$0.43}             \\
\bottomrule
\end{tabular}}
\label{tab_tau}
\end{table}

\begin{figure}[h]
\centering 
\includegraphics[width=0.8\textwidth]{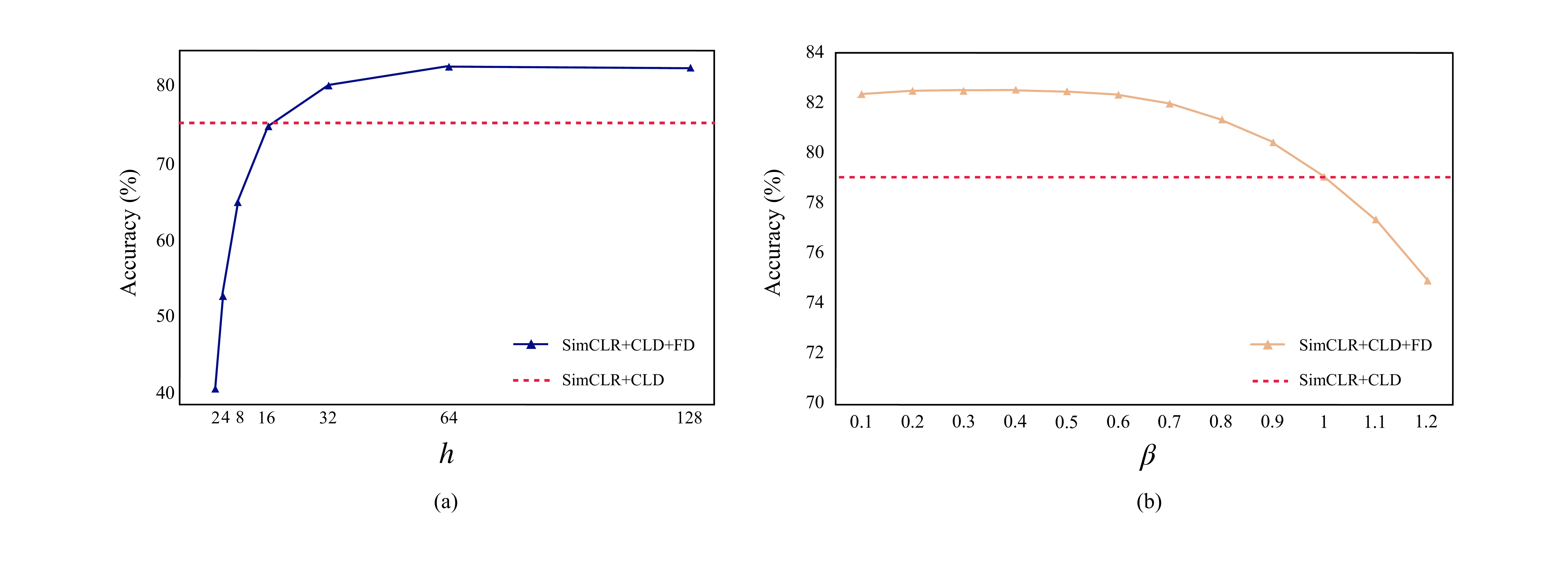} 
\caption{(a) Accuracy \emph{vs.} $h$ when $\beta=0.4$. (b) Accuracy \emph{vs.} $\beta$ when $h=64$.} 
\label{topk_beta}
\end{figure}

\begin{table}[ht]
\caption{Comparison of using FD or not on EuroSAT (5-way 1-shot).}
\resizebox{1\columnwidth}{!}{

\begin{tabular}{@{}cccccccc@{}}
\toprule
         & SimCLR & BYOL     & SimCLR+CLD & BYOL+CLD & one-to-all distillation  & Transfer+SimCLR & STARTUP \\ \hline
w/o FD   &75.10$\pm$0.85 &64.38$\pm$0.93 &79.38$\pm$0.82 &69.47$\pm$0.84   &77.70$\pm$0.83 &65.92$\pm$0.88 &65.24$\pm$0.88           \\
w/ FD     &78.02$\pm$0.77 &69.80$\pm$0.85            &82.52$\pm$0.76 &72.78$\pm$0.84 &80.62$\pm$0.77 &63.97$\pm$0.85            &60.87$\pm$0.88          \\ \midrule
$\Delta$ &+2.92        &+5.42            &+3.20          &+3.31            &+2.92  &-1.95 &-4.37       \\ \bottomrule
\end{tabular}
}
\label{tab_wwofd}
\end{table}

\subsection{Ablation Study on Feature Denoising}
On EuroSAT, the model SimCLR+CLD+FD is used to find optimal $h$ and $\beta$ on the 5-way 1-shot tasks. As shown in Figure \ref{topk_beta}, $h=64$ and $\beta=0.4$ are chosen for the best accuracy.

\begin{figure}[ht]

\begin{minipage}[]{0.47\linewidth}
\centering
\includegraphics[width=0.9\linewidth]{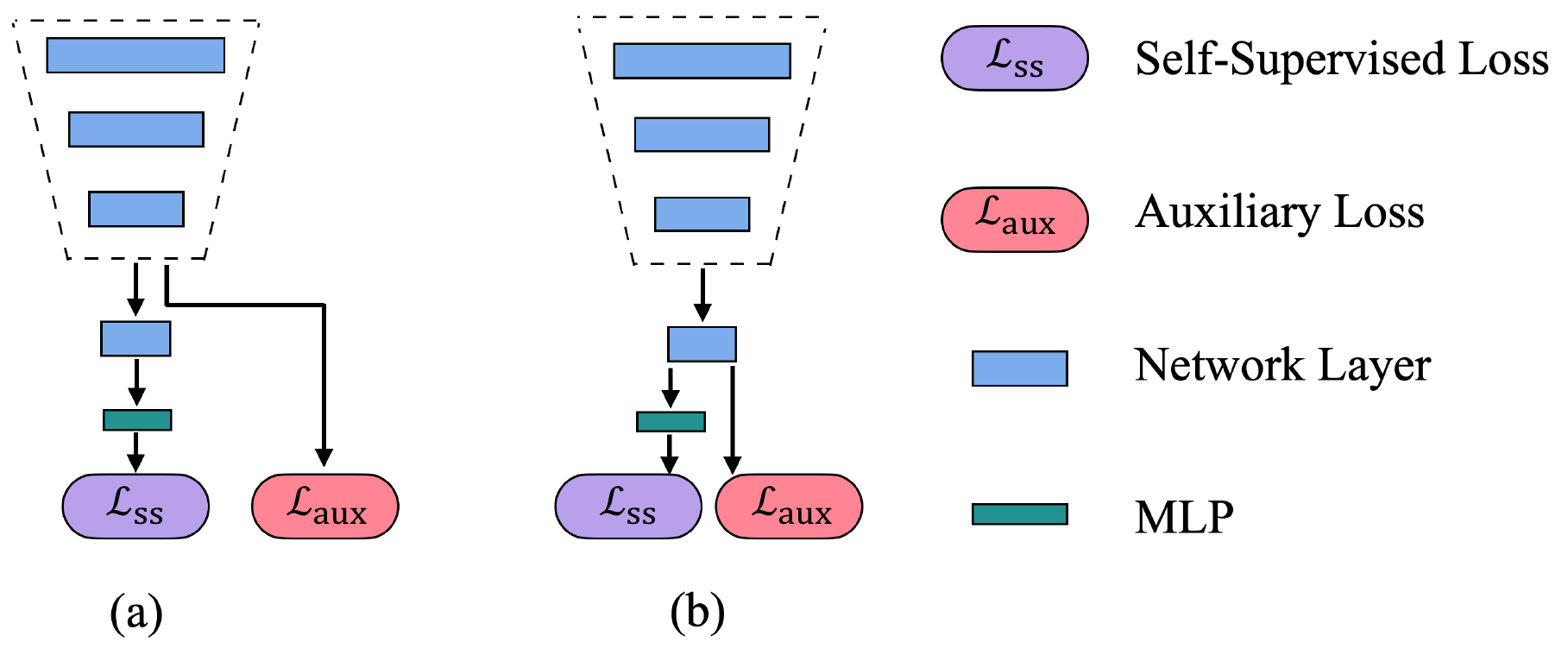}
\caption{$\mathcal{L}_{\rm{ss}}$ is the self-supervised loss and $\mathcal{L}_{\rm{aux}}$ includes other losses such as the cross-entropy and KD losses. (a) $\mathcal{L}_{\rm{aux}}$ is applied to intermediate layers. (b) $\mathcal{L}_{\rm{aux}}$ is applied to the final layer.} 
\label{fig_fd_works} 
\end{minipage}%
\hfill
\begin{minipage}[]{0.5\linewidth}
\centering
\includegraphics[width=0.9\linewidth]{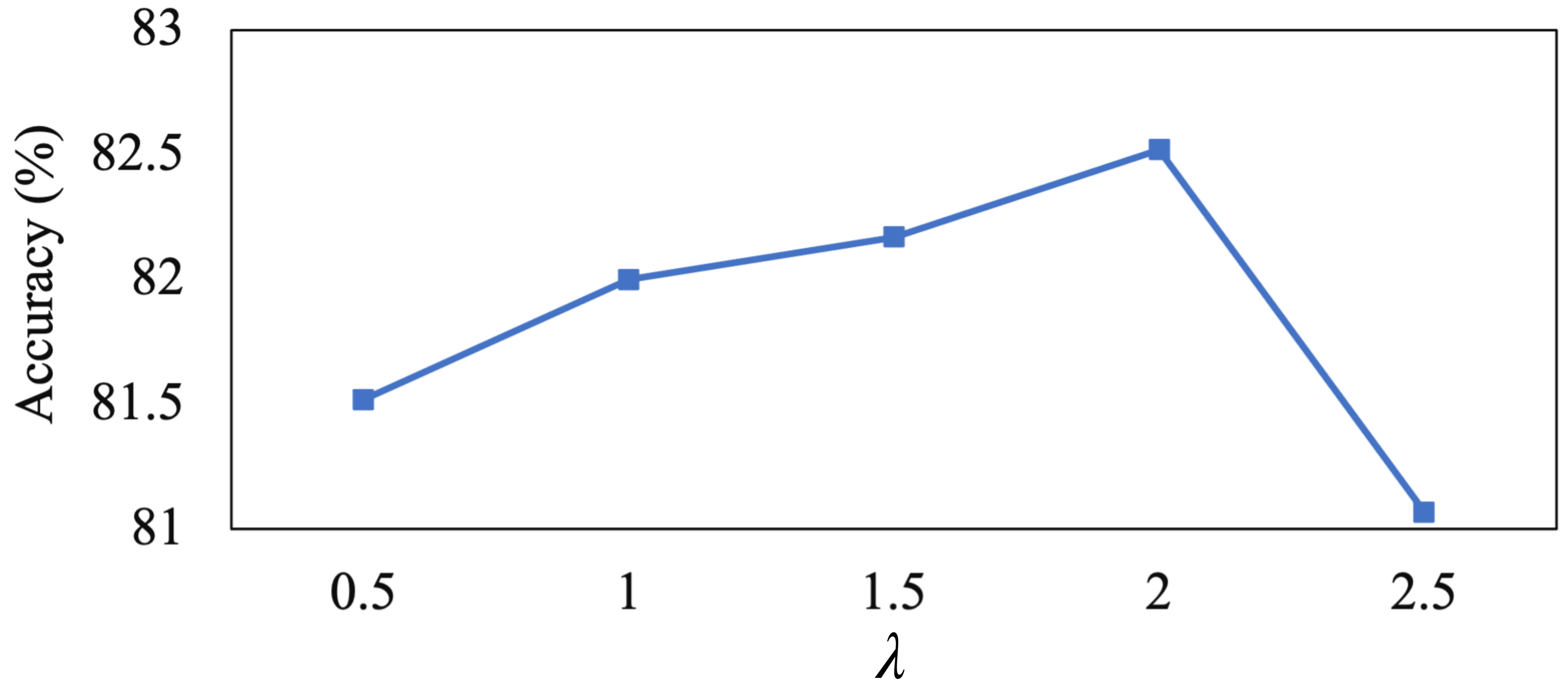}
\caption{Results of SimCLR+CLD+FD with the different $\lambda$ values on the EuroSAT dataset (5-way 1-shot).} 
\label{lambda}
\end{minipage}
\end{figure}

With $h=64$ and $\beta=0.4$ in FD, we compare using FD or not in Table \ref{tab_wwofd}. We can see that the first five models are improved with FD, while the last two are not. The different positions, which the auxiliary loss (cross-entropy and/or KD) is applied to in the two groups, are shown in Figures \ref{fig_fd_works}(a) and \ref{fig_fd_works}(b), respectively. The reason why FD does not help in Figure \ref{fig_fd_works}(b) is that the loss $\mathcal{L}_{\rm{aux}}$ used in the final layer reduces the noisy elements of the final feature vector for classification.

\subsection{Ablation Study on Weight Coefficient of Loss Function}

We give the results of SimCLR+CLD+FD with the different $\lambda$ values in Equation \ref{equ:loss_total} on the EuroSAT dataset (5-way 1-shot), as shown in Figure \ref{lambda}. We can see that the best result is obtained with $\lambda=2$. So we set $\lambda=2$ in all the related experiments.


\section{conclusion}

In this work, we handle the CD-FSC problems with large domain gaps between the base dataset and the target datasets. We propose the cross-level distillation KD for better transferring the knowledge of the base dataset to the student. We also present the feature denoising operation for mitigating the overfitting. Our method improves the performance of a strong baseline Dynamic-Distillation by 5.44$\%$ on 1-shot and 1.37$\%$ on 5-shot classification tasks on average in the BSCD-FSL benchmark, establishing new state-of-the-art results.

\section*{Acknowledgements}
We would also like to express our sincere gratitude to Mr. Haichen ZHENG, Mrs. Jing LIN for their unwavering support and confidence in our work.

\newpage

\bibliography{iclr2023_conference}

\begin{thebibliography}{30}
\providecommand{\natexlab}[1]{#1}
\providecommand{\url}[1]{\texttt{#1}}
\expandafter\ifx\csname urlstyle\endcsname\relax
  \providecommand{\doi}[1]{doi: #1}\else
  \providecommand{\doi}{doi: \begingroup \urlstyle{rm}\Url}\fi

\bibitem[Adler et~al.(2020)Adler, Brandstetter, Widrich, Mayr, Kreil, Kopp,
  Klambauer, and Hochreiter]{adler2020cross}
Thomas Adler, Johannes Brandstetter, Michael Widrich, Andreas Mayr, David
  Kreil, Michael Kopp, G{\"u}nter Klambauer, and Sepp Hochreiter.
\newblock Cross-domain few-shot learning by representation fusion.
\newblock \emph{arXiv:2010.06498}, 2020.

\bibitem[Chen et~al.(2020)Chen, Kornblith, Norouzi, and Hinton]{chen2020simple}
Ting Chen, Simon Kornblith, Mohammad Norouzi, and Geoffrey Hinton.
\newblock A simple framework for contrastive learning of visual
  representations.
\newblock In \emph{ICML}, 2020.

\bibitem[Chen et~al.(2018)Chen, Liu, Kira, Wang, and Huang]{chen2019closer}
Wei-Yu Chen, Yen-Cheng Liu, Zsolt Kira, Yu-Chiang~Frank Wang, and Jia-Bin
  Huang.
\newblock A closer look at few-shot classification.
\newblock In \emph{ICLR}, 2018.

\bibitem[Chen \& He(2021)Chen and He]{chen2021exploring}
Xinlei Chen and Kaiming He.
\newblock Exploring simple siamese representation learning.
\newblock In \emph{CVPR}, 2021.

\bibitem[Codella et~al.(2019)Codella, Rotemberg, Tschandl, Celebi, Dusza,
  Gutman, Helba, Kalloo, Liopyris, Marchetti, et~al.]{codella2019skin}
Noel Codella, Veronica Rotemberg, Philipp Tschandl, M~Emre Celebi, Stephen
  Dusza, David Gutman, Brian Helba, Aadi Kalloo, Konstantinos Liopyris, Michael
  Marchetti, et~al.
\newblock Skin lesion analysis toward melanoma detection 2018: A challenge
  hosted by the international skin imaging collaboration (isic).
\newblock \emph{arXiv:1902.03368}, 2019.

\bibitem[Das et~al.(2021)Das, Yun, and Porikli]{das2021confess}
Debasmit Das, Sungrack Yun, and Fatih Porikli.
\newblock Confess: A framework for single source cross-domain few-shot
  learning.
\newblock In \emph{ICLR}, 2021.

\bibitem[Fu et~al.(2021)Fu, Fu, and Jiang]{fu2021meta}
Yuqian Fu, Yanwei Fu, and Yu-Gang Jiang.
\newblock Meta-fdmixup: Cross-domain few-shot learning guided by labeled target
  data.
\newblock In \emph{ACM MM}, 2021.

\bibitem[Gidaris et~al.(2018)Gidaris, Singh, and
  Komodakis]{gidaris2018unsupervised}
Spyros Gidaris, Praveer Singh, and Nikos Komodakis.
\newblock Unsupervised representation learning by predicting image rotations.
\newblock In \emph{ICLR}, 2018.

\bibitem[Grill et~al.(2020)Grill, Strub, Altch{\'e}, Tallec, Richemond,
  Buchatskaya, Doersch, Avila~Pires, Guo, Gheshlaghi~Azar,
  et~al.]{grill2020bootstrap}
Jean-Bastien Grill, Florian Strub, Florent Altch{\'e}, Corentin Tallec, Pierre
  Richemond, Elena Buchatskaya, Carl Doersch, Bernardo Avila~Pires, Zhaohan
  Guo, Mohammad Gheshlaghi~Azar, et~al.
\newblock Bootstrap your own latent-a new approach to self-supervised learning.
\newblock In \emph{NeurIPS}, 2020.

\bibitem[Guo et~al.(2020)Guo, Codella, Karlinsky, Codella, Smith, Saenko,
  Rosing, and Feris]{guo2020broader}
Yunhui Guo, Noel~C Codella, Leonid Karlinsky, James~V Codella, John~R Smith,
  Kate Saenko, Tajana Rosing, and Rogerio Feris.
\newblock A broader study of cross-domain few-shot learning.
\newblock In \emph{ECCV}, 2020.

\bibitem[Helber et~al.(2019)Helber, Bischke, Dengel, and
  Borth]{helber2019eurosat}
Patrick Helber, Benjamin Bischke, Andreas Dengel, and Damian Borth.
\newblock Eurosat: A novel dataset and deep learning benchmark for land use and
  land cover classification.
\newblock \emph{IEEE Journal of Selected Topics in Applied Earth Observations
  and Remote Sensing}, pp.\  2217--2226, 2019.

\bibitem[Hinton et~al.(2015)Hinton, Vinyals, Dean,
  et~al.]{hinton2015distilling}
Geoffrey Hinton, Oriol Vinyals, Jeff Dean, et~al.
\newblock Distilling the knowledge in a neural network.
\newblock \emph{NeurIPS Workshop}, 2015.

\bibitem[Hua et~al.(2021)Hua, Wang, Xue, Ren, Wang, and Zhao]{hua2021feature}
Tianyu Hua, Wenxiao Wang, Zihui Xue, Sucheng Ren, Yue Wang, and Hang Zhao.
\newblock On feature decorrelation in self-supervised learning.
\newblock In \emph{ICCV}, 2021.

\bibitem[Islam et~al.(2021{\natexlab{a}})Islam, Chen, Panda, Karlinsky, Feris,
  and Radke]{islam2021dynamic}
Ashraful Islam, Chun-Fu~Richard Chen, Rameswar Panda, Leonid Karlinsky, Rogerio
  Feris, and Richard~J Radke.
\newblock Dynamic distillation network for cross-domain few-shot recognition
  with unlabeled data.
\newblock In \emph{NeurIPS}, 2021{\natexlab{a}}.

\bibitem[Islam et~al.(2021{\natexlab{b}})Islam, Chen, Panda, Karlinsky, Radke,
  and Feris]{islam2021broad}
Ashraful Islam, Chun-Fu~Richard Chen, Rameswar Panda, Leonid Karlinsky, Richard
  Radke, and Rogerio Feris.
\newblock A broad study on the transferability of visual representations with
  contrastive learning.
\newblock In \emph{ICCV}, 2021{\natexlab{b}}.

\bibitem[Kalibhat et~al.(2022)Kalibhat, Narang, Firooz, Sanjabi, and
  Feizi]{kalibhat2022towards}
Neha~Mukund Kalibhat, Kanika Narang, Hamed Firooz, Maziar Sanjabi, and Soheil
  Feizi.
\newblock Towards better understanding of self-supervised representations.
\newblock In \emph{ICML Workshop}, 2022.

\bibitem[Lu et~al.(2022)Lu, Wen, Liu, Liu, and Tian]{lu2022self}
Yuning Lu, Liangjian Wen, Jianzhuang Liu, Yajing Liu, and Xinmei Tian.
\newblock Self-supervision can be a good few-shot learner.
\newblock In \emph{ECCV}, 2022.

\bibitem[Luo et~al.(2022)Luo, Xu, and Xu]{luo2022channel}
Xu~Luo, Jing Xu, and Zenglin Xu.
\newblock Channel importance matters in few-shot image classification.
\newblock In \emph{ICML}, 2022.

\bibitem[Mangla et~al.(2020)Mangla, Kumari, Sinha, Singh, Krishnamurthy, and
  Balasubramanian]{mangla2020charting}
Puneet Mangla, Nupur Kumari, Abhishek Sinha, Mayank Singh, Balaji
  Krishnamurthy, and Vineeth~N Balasubramanian.
\newblock Charting the right manifold: Manifold mixup for few-shot learning.
\newblock In \emph{WACV}, 2020.

\bibitem[Mohanty et~al.(2016)Mohanty, Hughes, and
  Salath{\'e}]{mohanty2016using}
Sharada~P Mohanty, David~P Hughes, and Marcel Salath{\'e}.
\newblock Using deep learning for image-based plant disease detection.
\newblock \emph{Frontiers in plant science}, pp.\  1419--1432, 2016.

\bibitem[Phoo \& Hariharan(2021)Phoo and Hariharan]{phoo2020self}
Cheng~Perng Phoo and Bharath Hariharan.
\newblock Self-training for few-shot transfer across extreme task differences.
\newblock In \emph{ICLR}, 2021.

\bibitem[Rand(1971)]{rand1971objective}
William~M Rand.
\newblock Objective criteria for the evaluation of clustering methods.
\newblock \emph{Journal of the American Statistical association}, pp.\
  846--850, 1971.

\bibitem[Romero et~al.(2014)Romero, Ballas, Kahou, Chassang, Gatta, and
  Bengio]{romero2014fitnets}
Adriana Romero, Nicolas Ballas, Samira~Ebrahimi Kahou, Antoine Chassang, Carlo
  Gatta, and Yoshua Bengio.
\newblock Fitnets: Hints for thin deep nets.
\newblock \emph{arXiv:1412.6550}, 2014.

\bibitem[Triantafillou et~al.(2021)Triantafillou, Larochelle, Zemel, and
  Dumoulin]{triantafillou2021learning}
Eleni Triantafillou, Hugo Larochelle, Richard Zemel, and Vincent Dumoulin.
\newblock Learning a universal template for few-shot dataset generalization.
\newblock In \emph{ICML}, 2021.

\bibitem[Tseng et~al.(2020)Tseng, Lee, Huang, and Yang]{tseng2020cross}
Hung-Yu Tseng, Hsin-Ying Lee, Jia-Bin Huang, and Ming-Hsuan Yang.
\newblock Cross-domain few-shot classification via learned feature-wise
  transformation.
\newblock In \emph{ICLR}, 2020.

\bibitem[Vinyals et~al.(2016)Vinyals, Blundell, Lillicrap, Wierstra,
  et~al.]{vinyals2016matching}
Oriol Vinyals, Charles Blundell, Timothy Lillicrap, Daan Wierstra, et~al.
\newblock Matching networks for one shot learning.
\newblock In \emph{NeurIPS}, 2016.

\bibitem[Wang \& Deng(2021)Wang and Deng]{wang2021cross}
Haoqing Wang and Zhi-Hong Deng.
\newblock Cross-domain few-shot classification via adversarial task
  augmentation.
\newblock In \emph{IJCAI}, 2021.

\bibitem[Wang et~al.(2017)Wang, Peng, Lu, Lu, Bagheri, and
  Summers]{wang2017chestx}
Xiaosong Wang, Yifan Peng, Le~Lu, Zhiyong Lu, Mohammadhadi Bagheri, and
  Ronald~M Summers.
\newblock Chestx-ray8: Hospital-scale chest x-ray database and benchmarks on
  weakly-supervised classification and localization of common thorax diseases.
\newblock In \emph{CVPR}, 2017.

\bibitem[Zagoruyko \& Komodakis(2017)Zagoruyko and
  Komodakis]{zagoruyko2016paying}
Sergey Zagoruyko and Nikos Komodakis.
\newblock Paying more attention to attention: Improving the performance of
  convolutional neural networks via attention transfer.
\newblock In \emph{ICLR}, 2017.

\bibitem[Zhang et~al.(2019)Zhang, Song, Gao, Chen, Bao, and Ma]{zhang2019your}
Linfeng Zhang, Jiebo Song, Anni Gao, Jingwei Chen, Chenglong Bao, and Kaisheng
  Ma.
\newblock Be your own teacher: Improve the performance of convolutional neural
  networks via self distillation.
\newblock In \emph{ICCV}, 2019.

\end{thebibliography}
\bibliographystyle{iclr2023_conference}

\newpage

\appendix
\section{Appendix}
\label{sec:appendix}

\subsection{Results of the different batch sizes}
\begin{figure}[ht]
\centering 
\includegraphics[width=0.6\textwidth]{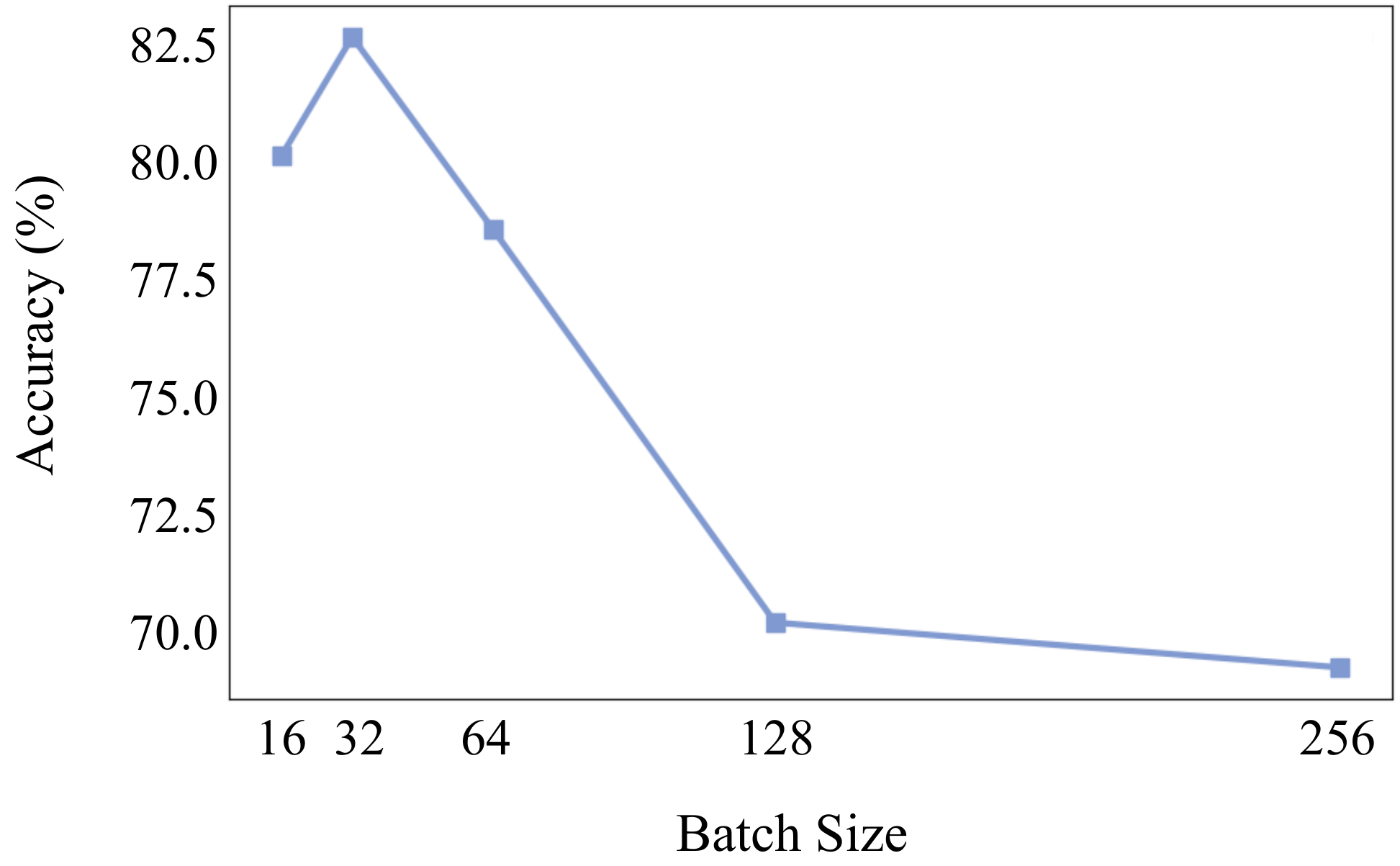} 
\caption{Results of the SimCLR+CLD+FD with the different batch sizes on the EuroSAT dataset (5-way 1-shot).} 
\label{batch_size}
\end{figure}

We present the results of the SimCLR+CLD+FD trained with different batch sizes on the EuroSAT dataset (5-way 1-shot) in Figure \ref{batch_size}. We can see that the optimal batch size is 32, so we utilize this value in all the experiments of our method.

\subsection{Results of the different percentages of unlabeled images}

\begin{figure}[ht]
\centering 
\includegraphics[width=0.6\textwidth]{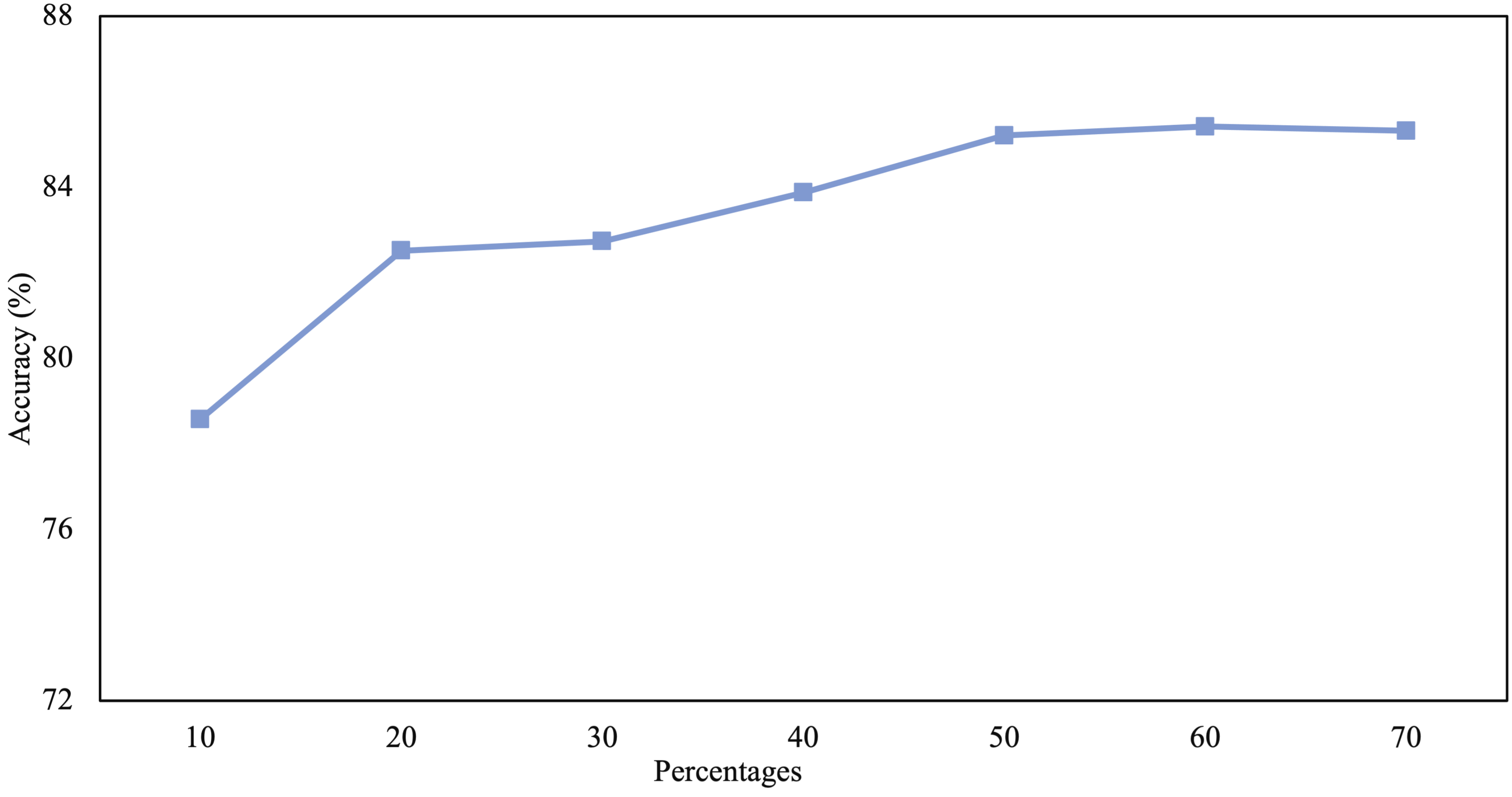} 
\caption{Results of the SimCLR+CLD+FD with the different percentages on the EuroSAT dataset (5-way 1-shot).} 
\label{percentage}
\end{figure}

Figure \ref{percentage} shows that as more unlabeled target images become available, the accuracy of the model will gradually converge to a plateau.

\subsection{Results with self-supervised pre-trained teacher}

\begin{table}[ht]
\caption{Results with self-supervised \& supervised pre-trained teacher.}
\resizebox{1\columnwidth}{!}
{
\begin{tabular}{@{}ccccccccc@{}}
\toprule
                & \multicolumn{2}{c}{EuroSAT} & \multicolumn{2}{c}{CropDisease} & \multicolumn{2}{c}{ISIC} & \multicolumn{2}{c}{ChestX} \\
                & 1-shot       & 5-shot       & 1-shot         & 5-shot         & 1-shot      & 5-shot     & 1-shot       & 5-shot      \\ \midrule
SimCLR+CLR+FD (self-supervised) &80.30$\pm$0.72 &91.63$\pm$0.36 &89.94$\pm$0.72 &96.50$\pm$0.35 &37.42$\pm$0.46 &49.36$\pm$0.64 &22.58$\pm$0.43 &25.96$\pm$0.43  \\
SimCLR+CLR+FD (supervised) &82.52$\pm$0.76 &92.89$\pm$0.34 &90.48$\pm$0.72 &96.58$\pm$0.39 &39.70$\pm$0.69 &52.29$\pm$0.62 &22.39$\pm$0.44  &25.98$\pm$0.43 \\

\bottomrule
\end{tabular}}
\label{tab_selfteacher}
\end{table}

On the EuroSAT dataset (5-way 1-shot), we give the results of the SimCLR+FD+CLD where the teacher is pre-trained in a self-supervised scheme. The performances of the SimCLR+CLD+FD are shown in Table \ref{tab_selfteacher}.

\end{document}